\documentclass{article}
\usepackage{spconf,amsmath,amssymb,graphicx}
\usepackage{booktabs,multirow}
\usepackage[table]{xcolor}
\usepackage{algorithm,algorithmic}
\usepackage{enumitem}
\usepackage{pifont}
\usepackage{tikz}

\newlength{\barW}
\newcommand{\abar}[5]{\begin{tikzpicture}[baseline=0.1ex,x=\barW]
  \useasboundingbox (0,0) rectangle (1,0.17);
  \fill[#5] ({(#2-#3)/(#4-#3)},0.035) rectangle ({(#1-#3)/(#4-#3)},0.155);
  \draw[jmGrey,line width=0.35pt] ({(#2-#3)/(#4-#3)},-0.02) -- ++(0,0.21);
\end{tikzpicture}}
\newcommand{\nab}{\hphantom{0.00}\na\,\hphantom{\rule{\barW}{1pt}}}   
\newcommand{\tb}[2]{#1\,\abar{#1}{0.3333}{0.27}{0.57}{#2}}   
\newcommand{\bb}[2]{#1\,\abar{#1}{0.5}{0.42}{0.78}{#2}}      
\newcommand{\ab}[2]{#1\,\abar{#1}{0.5}{0.42}{0.82}{#2}}      
\newcommand{\abest}[2]{\best{#1}\,\abar{#1}{0.5}{0.42}{0.82}{#2}}
\newcommand{\tbest}[2]{\best{#1}\,\abar{#1}{0.3333}{0.27}{0.57}{#2}}
\newcommand{\bbest}[2]{\best{#1}\,\abar{#1}{0.5}{0.42}{0.78}{#2}}

\setlist{nosep, leftmargin=14pt}
\AtBeginDocument{%
  \setlength{\abovedisplayskip}{4pt plus 1pt minus 1pt}%
  \setlength{\belowdisplayskip}{4pt plus 1pt minus 1pt}%
  \setlength{\abovedisplayshortskip}{2pt plus 1pt}%
  \setlength{\belowdisplayshortskip}{2pt plus 1pt}}

\definecolor{jmRed}{HTML}{B8432F}
\definecolor{jmBlue}{HTML}{3F5F8A}
\definecolor{jmGrey}{HTML}{9A9A9A}
\definecolor{jmInk}{HTML}{222222}
\definecolor{jmTint}{HTML}{F7ECE8}
\newcommand{\cmark}{\textcolor{jmBlue}{\ding{51}}}
\newcommand{\xmark}{\textcolor{jmGrey}{\ding{55}}}
\newcommand{\na}{\textcolor{jmGrey}{--}}
\newcommand{\best}[1]{\textbf{\textcolor{jmRed}{#1}}}
\colorlet{cOurs}{jmRed}\colorlet{cLin}{jmRed!55}\colorlet{cCls}{jmRed!30}
\colorlet{cFeat}{jmBlue!55}\colorlet{cRaw}{jmGrey!70}

\newcommand{\R}{\mathbb{R}}
\newcommand{\Spd}[1]{\mathcal{S}_{++}^{#1}}

\newcommand{\vecu}{\operatorname{vec}_{\mathcal{T}}}
\newcommand{\vech}{\operatorname{vech}}
\newcommand{\norm}[1]{\lVert #1 \rVert}
\newcommand{\Itr}{\mathcal{I}_{\mathrm{tr}}}
\newcommand{\Ite}{\mathcal{I}_{\mathrm{te}}}

\newcommand{\inner}[2]{\langle #1, #2 \rangle}

\title{ACAM-J-MARS: Multi-Atlas Riemannian Spectral Representations\\for Decoding Advanced Meditation States from fMRI}

\name{Jian Wang$^{1,2}$, Winson F.~Z. Yang$^{1,2}$, Wugeng Zheng$^{3}$, Matthew D. Sacchet$^{1,2}$}
\address{$^{1}$Meditation Research Program, Psychiatry, MGH and Harvard Medical School, Boston, MA, USA\\
$^{2}$Martinos Center for Biomedical Imaging, Radiology, MGH and Harvard Medical School, Boston, MA, USA\\
$^{3}$College of Engineering, Northeastern University, Boston, MA, USA}

\begin{document}
\renewcommand{\thefootnote}{\fnsymbol{footnote}}
\maketitle
\renewcommand{\thefootnote}{\arabic{footnote}}\setcounter{footnote}{0}

\begin{abstract}
Inferring jhana-type advanced concentration absorption meditation (ACAM-J) states from fMRI data is challenging due to the spatially distributed nature of meditation-related effects, substantial inter-individual variability that can exceed within-subject state differences, and limited sample sizes. Conventional machine-learning approaches often formulate this problem as unimodal, Euclidean, atlas-based classification, thereby neglecting the non-Euclidean geometry of functional connectivity matrices and underrepresenting temporal information in functional magnetic resonance imaging (fMRI); these limitations can lead to suboptimal classification performance. We propose ACAM-J-MARS (\textbf{M}ulti-\textbf{A}tlas \textbf{R}iemannian \textbf{S}pectral representation), which computes functional connectivity using multiple atlases, represents symmetric positive definite (SPD) connectivity matrices on a Riemannian manifold, projects them into a log-Euclidean tangent space, and augments connectivity-derived features with low-frequency BOLD spectral energy to better resolve subtle state differences under heterogeneous variability. Evaluated on 204 segments from 22 practitioners, ACAM-J-MARS improves performance across multiple backbone models, achieving a balanced accuracy of 0.757 for early versus late ACAM-J classification and 0.545 for three-class classification, corresponding to a 10.8-point absolute improvement over the strongest baseline. These findings suggest that the proposed representation improves the decoding of advanced meditative brain states and may enable downstream applications such as closed-loop neurofeedback and real-time psychophysiological brain-computer interfaces.\footnote{Source code available upon request: jwang186@mgh.harvard.edu}
\end{abstract}


\section{INTRODUCTION}
\label{sec:intro}
\begin{figure}[!tbh]
\centering
\includegraphics[width=\columnwidth]{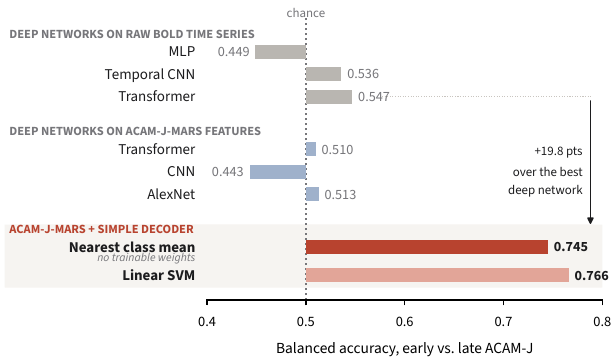}
\caption{\textbf{Deep neural networks cannot decode ACAM-J states across subjects.} Under subject-wise cross-validation, deep models are near chance (0.44--0.55) on raw BOLD or ACAM-J-MARS features, whereas simpler classifiers perform well (nearest class mean, 0.745; linear SVM, 0.766).}
\label{fig:teaser}
\end{figure}
Meditation offers a controlled way to study the neural regulation of attention, arousal, and self-related processing \cite{sacchet2026tics,sacchet2026neuron}, and different practices recruit partly distinct brain systems \cite{baten2026functional}. Jhana, or advanced concentration absorption meditation (ACAM-J), is a sequence of eight progressive absorption states (ACAM-J1--J8). With increasing absorption, practitioners report steadier attention, reduced sensory awareness, and quieter discursive thought \cite{sparby2024unified,yang2024volitional}. Advanced meditators can enter ACAM-J volitionally during scanning \cite{yang2024jhana,yang2025neuroscience}, making ACAM-J a rare model of internally generated, repeatable brain-state changes. 7T fMRI case studies of a single advanced meditator have linked ACAM-J to dynamic brain states \cite{treves2025dynamic}, connectivity gradients \cite{demir2025gradients}, cortical eigenmodes \cite{potash2025eigenmode}, and network modularity \cite{chowdhury2025multimodal}, and related work points to changes in nonlinear processing and near-critical whole-brain dynamics \cite{potash2025integrated,vohryzek2025whole}. However, because most evidence comes from one person, it is unclear whether ACAM-J states have a shared neural signature across practitioners. A prior cross-subject classifier using local voxel synchrony achieved ${\sim}66\%$ accuracy in distinguishing state pairs in one held-out meditator \cite{kumar2026ml}.
A decoder of ACAM-J states that generalizes across people would turn single-case findings into a population-level account, but three challenges hinder this. First, changes across ACAM-J states likely reflect distributed cortical–subcortical interactions for attention and arousal \cite{halassa2017thalamic}, not focal activations. Second, functional connectivity is strongly individual \cite{finn2015functional} and can dominate transient state changes \cite{gratton2018functional}, so models trained to minimize average error may learn subject identity rather than ACAM-J state. Third, whole-brain connectivity is high-dimensional relative to the data (204 segments, 22 practitioners), making overfitting likely, and end-to-end deep networks offer little advantage in this setting \cite{he2020deep}.

Most connectivity decoders compute Pearson correlations within a cortical atlas and vectorize the matrix in Euclidean space \cite{dadi2019benchmarking}, losing key structure: the non-Euclidean geometry of SPD correlation matrices \cite{pennec2006riemannian}, contributions from subcortical and brainstem regions \cite{bianciardi2015toward}, and temporal dynamics. Bigger models cannot recover this lost information: multilayer perceptrons, CNNs, and Transformers stay near chance on raw BOLD or derived features, while simple classifiers on the same features perform well (Fig.~\ref{fig:teaser}).
\begin{figure*}[!thb]
\centering
\includegraphics[width=\linewidth]{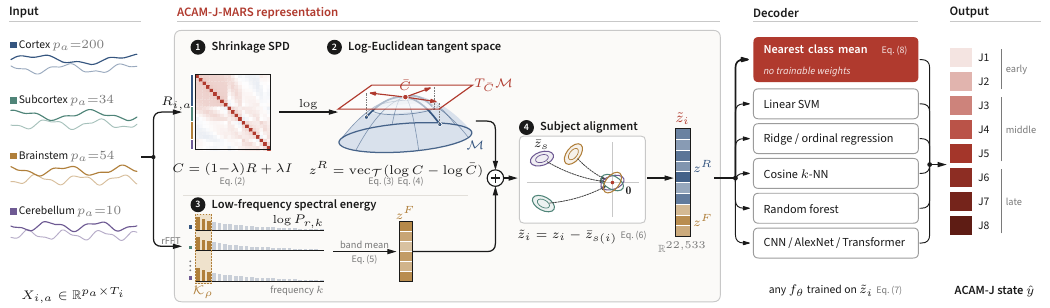}
\caption{\textbf{ACAM-J-MARS.} Regional BOLD signals from four atlases are converted to shrinkage-regularised SPD connectivity matrices, mapped to the log-Euclidean tangent space at the training-set Fr\'echet mean, and concatenated with low-frequency spectral energy. After per-subject mean removal, the resulting features can be decoded by classifiers ranging from nearest class-mean to deep neural networks.}
\label{fig:overview}
\end{figure*}

To address these challenges, we propose ACAM-J-MARS with the main contributions as follows:

\begin{itemize}[leftmargin=*]
    \item \textbf{Unified geometrical–spectral representation:} We introduce \textbf{ACAM-J-MARS}, which fuses multi-region connectivity (cortical, subcortical, brainstem, cerebellar) in the log-Euclidean tangent space of SPD matrices with regional low-frequency spectral energy. This shared vector space reduces inter-subject variability through translation adjustment, avoiding labeled calibration data.
    
    \item \textbf{Classifier-agnostic decoding gains:} The feature space supports robusst decoding. A parameter-free nearest class-mean classifier achieves 0.757 balanced accuracy for early vs.\ late ACAM-J (\textbf{+10.8\%} over the best baseline). Five other linear/centroid classifiers score 0.70--0.77 on the same features.
    
    \item \textbf{Objective absorption biomarker:} Across 22 practitioners, we identify a cross-subject neural correlate of absorption (ACAM-J) states. This objective measure complements introspection and supports real-time monitoring and closed-loop neurofeedback.
\end{itemize}
\begin{table*}[t]
\centering
\caption{\textbf{Subject-level cross-validation for ACAM-J decoding.} Rows cumulatively add ACAM-J-MARS components. Atlases: S200 = Schaefer200; All = $\mathcal{A}$ (four parcellations). Geometry: Euc. = Euclidean vectorisation; LE = log-Euclidean tangent space. Classifiers: NCM = nearest class mean (cosine); Ridge = $\ell_2$-regularised linear regression. Chance: 0.500 (two-class), 0.333 (three-class). \na\ = not tested. Best in red. Right: stage-specific regional differences vs.\ ACAM-J1 (MNI $z{=}24$\,mm; row-wise scale).}
\label{tab:main_results}
\renewcommand{\arraystretch}{1.12}
\footnotesize
\setlength{\tabcolsep}{2.4pt}
\begin{minipage}[c]{0.635\textwidth}\centering\resizebox{\linewidth}{!}{%
\begin{tabular}{@{}l cccc c ccc c ccc@{}}
\toprule
& \multicolumn{4}{c}{Representation} && \multicolumn{3}{c}{Binary (early vs.\ late)} && \multicolumn{3}{c}{Ternary (3-level)} \\
\cmidrule{2-5}\cmidrule{7-9}\cmidrule{11-13}
Method & Atlas & Geometry & Spectral & Classifier && Acc. & BA & F1 && Acc. & BA & F1 \\
\midrule
\multicolumn{13}{@{}l}{\textit{\textcolor{jmGrey}{Unimodal Euclidean connectivity}}} \\
\quad S200 FC & S200 & Euc. & \xmark & NCM && 0.576 & 0.579 & 0.574 && \na & \na & \na \\
\quad S200 FC & S200 & Euc. & \xmark & Ridge && 0.668 & 0.649 & 0.645 && \na & \na & \na \\
\quad All-atlas FC & All & Euc. & \xmark & Ridge && 0.659 & 0.641 & 0.637 && \na & \na & \na \\
\addlinespace[2pt]
\multicolumn{13}{@{}l}{\textit{\textcolor{jmBlue}{Partial geometric representations}}} \\
\quad Spectral only & All & \na & \cmark & NCM && \na & \na & \na && 0.381 & 0.405 & 0.352 \\
\quad Tangent only & All & LE & \xmark & NCM && 0.695 & 0.696 & 0.692 && 0.472 & 0.485 & 0.472 \\
\quad S200 tangent + spec. & S200 & LE & \cmark & NCM && \na & \na & \na && 0.495 & 0.515 & 0.497 \\
\addlinespace[2pt]
\rowcolor{jmTint}
\textbf{ACAM-J-MARS} (ours) & All & LE & \cmark & NCM && \best{0.751} & \best{0.757} & \best{0.750} && 0.529 & 0.545 & 0.529 \\
\rowcolor{jmTint}
\quad + stage-level means & All & LE & \cmark & NCM && \na & \na & \na && \best{0.530} & \best{0.552} & \best{0.531} \\
\bottomrule
\end{tabular}}
\end{minipage}\hfill
\begin{minipage}[c]{0.325\textwidth}\centering
\includegraphics[width=\linewidth]{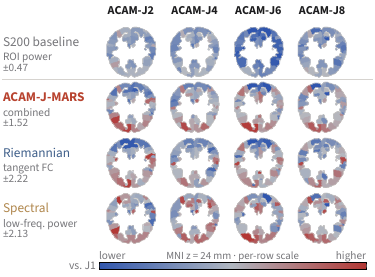}
\end{minipage}
\end{table*}

\section{UNIMODAL CONNECTIVITY AND ITS LIMITATIONS}
\label{sec:background}

Let $X_i \in \R^{p \times T_i}$ be the regional BOLD time series of run $i$, with ACAM-J state label $y_i$. A unimodal baseline computes the Pearson correlation matrix $R_i$ on one atlas~\cite{dadi2019benchmarking} and uses its upper triangle
\begin{equation}
z^{U}_i = \vech(R_i) \in \R^{p(p-1)/2}.
\label{eq:unimodal}
\end{equation}
This feature has four limitations: (1) $R_i\in\Spd{p}$, but the Euclidean distance $\|R_1-R_2\|_F$ ignores manifold curvature, and Euclidean averaging inflates determinants \cite{pennec2006riemannian,arsigny2006log}. (2) Cortical atlases omit subcortical and brainstem regions important for arousal and attention \cite{bianciardi2015toward,halassa2017thalamic}. (3) Correlation is order-invariant, so temporal dynamics are lost. (4) Subject offsets $\beta_{s(i)}$ add to the state signal and can dominate Euclidean distances \cite{gratton2018functional}. On our data, $z^U_i$ achieves 0.579 balanced accuracy with nearest class mean and 0.649 with ridge regression.

\section{METHOD: ACAM-J-MARS}
\label{sec:method}

ACAM-J-MARS projects BOLD time-series data into a subject-aligned feature vector (Fig.~\ref{fig:overview}). Each atlas $a \in \mathcal{A}$ comprises $p_a$ distinct regions.

\subsection{Tangent-Space Connectivity}
When \(T_i\) is close to \(p_a\), the sample correlation matrix \(R_{i,a}\) may become ill-conditioned or rank-deficient. To regularize this estimate, we apply shrinkage toward the identity matrix, corresponding to a simple Ledoit--Wolf-type estimator \cite{ledoit2004well}:
\begin{equation}
C_{i,a} = (1-\lambda)R_{i,a} + \lambda I_{p_a},\qquad \lambda \in (0,1).
\label{eq:shrink}
\end{equation}
Subsequent computations are performed on \(\Spd{p_a}\) endowed with the log-Euclidean metric \cite{arsigny2006log}. As a reference element, we use the log-Euclidean Fr\'echet mean of the training runs \cite{pennec2006riemannian},
\begin{equation}
\bar{C}_a = \exp\Big(\tfrac{1}{|\Itr|}\textstyle\sum_{i\in\Itr}\log C_{i,a}\Big),
\label{eq:frechet}
\end{equation}
and represent each run by its tangent-space coordinate at \(\bar C_a\) \cite{barachant2012multiclass}:
\begin{equation}
z^R_{i,a} = \vecu\big(\log C_{i,a} - \log\bar{C}_a\big) \in \R^{p_a(p_a+1)/2}.
\label{eq:tangent}
\end{equation}
Here, \(\log\) is the matrix logarithm. The operator \(\vecu\) vectorizes a matrix’s upper triangle and rescales off-diagonal entries by \(\sqrt{2}\) to preserve the Frobenius inner product \cite{barachant2012multiclass}~\footnote{The final connectivity representation is obtained by concatenating atlas-specific feature vectors, defined as \(z^R_i = \oplus_{a \in \mathcal{A}} z^R_{i,a}\).}.

\subsection{Low-Frequency Spectral Energy}
Because correlation-based measures ignore temporal ordering, we augment the representation with a single spectral descriptor for each region. Let $P_{i,a,r}(k)$ denote the periodogram of region $r$, and let $\mathcal{K}_\rho$ be the set of the lowest $\rho=10\%$ non-DC frequency bins. Following~\cite{zang2007altered}, we define
\begin{equation}
z^F_{i,a,r} = \tfrac{1}{|\mathcal{K}_\rho|}\textstyle\sum_{k\in\mathcal{K}_\rho}\log\big(1+P_{i,a,r}(k)\big).
\label{eq:fourier}
\end{equation}
The logarithmic transform is applied to stabilise variance and to convert multiplicative scaling effects into additive offsets. The resulting feature vector is formed by concatenating regional and spectral components, $z_i = z^R_i \oplus z^F_i$.

\subsection{Subject Alignment and Decoding}
We use the generative model $z_i=\beta_{s(i)}+\mu_{y_i}+\varepsilon_i$. Since the tangent space is a vector space, we remove the subject offset $\beta_{s(i)}$ by within-subject mean-centering~\cite{zanini2018transfer}:
\begin{equation}
\tilde{z}_i = z_i - \tfrac{1}{|\mathcal{I}_{s(i)}|}\textstyle\sum_{j\in\mathcal{I}_{s(i)}}z_j.
\label{eq:centre}
\end{equation}
This centering is unsupervised and can be applied to unseen subjects, provided multiple runs are available. A decoder $f_\theta$ is then trained on $\tilde z_i$ via ERM:
\begin{equation}
\hat{\theta} = \arg\min_\theta \textstyle\sum_{i \in \Itr} \mathcal{L}\big(f_\theta(\tilde{z}_i), y_i\big).
\label{eq:erm}
\end{equation}
To assess representations with less dependence on the decoder, we mainly report a nearest class-mean (NCM) classifier. Let $\hat{\mu}_c$ be the class mean on $\Itr$; we predict the class with highest cosine similarity:
\begin{equation}
\hat{y}_j = \arg\max_{c}\ \frac{\inner{\tilde{z}_j}{\hat{\mu}_c}}{\norm{\tilde{z}_j}_2 \norm{\hat{\mu}_c}_2}.
\label{eq:proto}
\end{equation}

\section{EXPERIMENTS}

\noindent\textbf{Data and Tasks.} We use 204 fMRI ACAM-J segments from 22 advanced meditators, each with a self-reported ACAM-J1--J8 label. The atlas set $\mathcal{A}$ includes Schaefer200 (cortex), Tian-S2 (subcortex), and brainstem and cerebellar atlases with $p_a \in \{200,34,54,10\}$ \cite{schaefer2018local,tian2020topographic,bianciardi2015toward,king2019functional}, yielding $d=22{,}533$ features. Main tasks are early (J1--J2) vs.\ late (J6--J8) and 3-class early/middle (J3--J5)/late. We also evaluate a median split (J1--J4 vs.\ J5--J8) and J1 vs.\ J8.

\noindent\textbf{Protocol \& Comparisons.} We use subject-wise cross-validation~\cite{varoquaux2017assessing} with disjoint subjects ($\Itr \cap \Ite = \emptyset$). Fr\'echet means, standardisation, and class means are computed on $\Itr$ only. We report balanced accuracy (BA), plus accuracy, AUC, and macro-F1 for additional contrasts. We compare: (1) single-atlas Euclidean FC with NCM and ridge; (2) multi-atlas FC in Euclidean and tangent space; (3) spectral vs.\ tangent features alone; and (4) alternative decoders on ACAM-J-MARS features (1D CNN, reduced AlexNet), plus deep networks on raw BOLD. To relate features to anatomy, we map each feature block’s stage-wise change relative to ACAM-J1 to brain regions (Table~\ref{tab:main_results}, right).

\section{RESULTS}
\label{sec:results}

\begin{figure}[!t]
\centering
\includegraphics[width=.95\columnwidth]{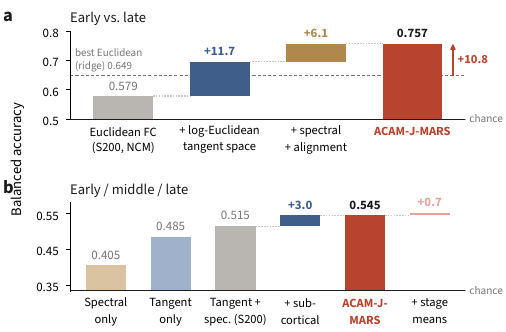}
\caption{\textbf{Component contributions.} Subject-wise cross-validated balanced accuracy. \textbf{a} Binary (early vs.\ late): log-Euclidean tangent features beat Euclidean FC by 11.7 points. \textbf{b} Ternary: spectral+tangent features combine; extra atlases help only in tangent space.}
\label{fig:ladder}
\end{figure}

Fig.~\ref{fig:teaser} and Tab.~\ref{tab:deep_results} show that deep neural network classifiers perform near chance when trained on raw BOLD time series or ACAM-J-MARS features, whereas linear and centroid-based decoders reach much higher accuracy (0.65--0.77). This likely reflects (i) the lack of local spatial structure in tangent-space coordinates exploited by CNNs, and (ii) overfitting of flexible non-linear models in high-dimensional, low-sample settings (d = 22{,}533; N = 204) \cite{he2020deep}.

Tab.~\ref{tab:main_results} and Fig.~\ref{fig:ladder} show that single-atlas Euclidean FC yields 0.579 (NCM) and 0.649 (ridge) for binary classification. Adding subcortical atlases increases accuracy to 0.641. Four-atlas tangent-space FC improves to 0.696, and adding spectral features plus subject-level alignment reaches 0.757 (+10.8 points vs.\ Euclidean baselines).

In Fig.~\ref{fig:ladder}b, for ternary classification, spectral features score 0.405 and tangent-space features 0.485. Fusing both with Schaefer200 increases accuracy to 0.515, and fusing across all four atlases to 0.545. Baseline maps from Schaefer200 show broadly uniform decreases, while ACAM-J-MARS maps show anterior--posterior gradients.

Tab.~\ref{tab:contrasts} shows Euclidean FC is near chance on the median split and ACAM-J1 vs.\ ACAM-J8 (AUC = 0.53, 0.47), while both versions of our representation are well above chance (AUC = 0.69--0.78). In both contrasts, the tangent-space component alone matches full ACAM-J-MARS (AUC = 0.70 vs.\ 0.69; 0.75 vs.\ 0.77). Spectral and alignment features drive early vs.\ late ACAM-J performance (BA 0.696 to 0.757, Table~\ref{tab:main_results}), with the full representation reaching AUC = 0.79--0.81.

\begin{table}[!t]
\centering
\caption{\textbf{Classifier Capacity and ACAM-J-MARS Features.} Deep networks: seed-averaged means. Chance is 0.333 (ternary) and 0.500 (binary).}
\label{tab:deep_results}
\renewcommand{\arraystretch}{0.94}
\footnotesize
\setlength{\tabcolsep}{4pt}
\begin{tabular}{@{}l r r@{}}
\toprule
Decoder & Ternary BA & Binary BA \\
\midrule
\multicolumn{3}{@{}l}{\textit{\textcolor{jmRed}{ACAM-J-MARS, centroid and linear decoders}}} \\
\rowcolor{jmTint}
\quad NCM, correlation (ours) & \tb{0.545}{cOurs} & \bb{0.757}{cOurs} \\
\quad Stage centroids$^{\dagger}$ & \tbest{0.552}{cLin} & \bb{0.691}{cLin} \\
\quad Linear SVM & \tb{0.519}{cLin} & \bbest{0.766}{cLin} \\
\quad Subject-rank ridge & \nab & \bb{0.744}{cLin} \\
\quad Ordinal regression & \nab & \bb{0.721}{cLin} \\
\quad Subject-rank endpoint axis & \tb{0.480}{cLin} & \bb{0.702}{cLin} \\
\quad Ridge & \tb{0.439}{cLin} & \bb{0.648}{cLin} \\
\addlinespace[1pt]
\multicolumn{3}{@{}l}{\textit{\textcolor{jmRed!80!black}{ACAM-J-MARS, non-parametric and non-linear}}} \\
\quad Cosine $k$-NN ($k{=}5$\,/\,$3$) & \tb{0.478}{cCls} & \bb{0.615}{cCls} \\
\quad Cosine 1-NN & \nab & \bb{0.594}{cCls} \\
\quad Random forest & \tb{0.332}{cCls} & \bb{0.511}{cCls} \\
\addlinespace[1pt]
\multicolumn{3}{@{}l}{\textit{\textcolor{jmBlue}{Deep networks, ACAM-J-MARS features}}} \\
\quad Transformer & \tb{0.362}{cFeat} & \bb{0.510}{cFeat} \\
\quad CNN & \tb{0.351}{cFeat} & \bb{0.443}{cFeat} \\
\quad AlexNet & \tb{0.339}{cFeat} & \bb{0.513}{cFeat} \\
\addlinespace[1pt]
\multicolumn{3}{@{}l}{\textit{\textcolor{jmGrey}{Deep networks, raw BOLD time series}}} \\
\quad MLP & \tb{0.293}{cRaw} & \bb{0.449}{cRaw} \\
\quad Temporal CNN & \tb{0.344}{cRaw} & \bb{0.536}{cRaw} \\
\quad Transformer & \tb{0.333}{cRaw} & \bb{0.547}{cRaw} \\
\bottomrule
\end{tabular}

\vspace{2pt}\parbox{\columnwidth}{\scriptsize\raggedright $^{\dagger}$Mean rule (ternary), max rule (binary). Bars start at chance (grey tick).}
\end{table}

\begin{table}[!t]
\caption{\textbf{Robustness to ACAM-J contrasts.} Subject-wise cross-validation. S200 FC: single-atlas Euclidean baseline; Ours (tangent): four-atlas log-Euclidean connectivity only; Ours (full): full ACAM-J-MARS with spectral energy and subject alignment. NCM: nearest class mean; SVM: linear support vector machine. Bars show AUC above chance (0.5).}
\label{tab:contrasts}
\renewcommand{\arraystretch}{0.94}
\footnotesize
\setlength{\tabcolsep}{2.6pt}
\centering
\begin{tabular}{@{}l l l c r c@{}}
\toprule
Contrast & Representation & Dec. & Acc. & \multicolumn{1}{c}{AUC} & F1 \\
\midrule
ACAM-J1--J2 & Ours (full) & NCM & 0.740 & \ab{0.788}{cOurs} & 0.739 \\
\quad vs.\ ACAM-J6--J8 & Ours (full) & SVM & \best{0.764} & \abest{0.805}{cOurs} & \best{0.762} \\
\addlinespace[1pt]
ACAM-J1--J4 & S200 FC & NCM & 0.554 & \ab{0.530}{cRaw} & 0.551 \\
\quad vs.\ ACAM-J5--J8 & Ours (tangent) & NCM & \best{0.681} & \abest{0.702}{cLin} & \best{0.681} \\
                   & Ours (full) & NCM & 0.652 & \ab{0.688}{cOurs} & 0.652 \\
\addlinespace[1pt]
ACAM-J1 & S200 FC & NCM & 0.451 & \ab{0.470}{cRaw} & 0.451 \\
\quad vs.\ ACAM-J8 & Ours (tangent) & NCM & 0.725 & \ab{0.753}{cLin} & 0.725 \\
                   & Ours (tangent) & SVM & \best{0.784} & \ab{0.775}{cLin} & \best{0.781} \\
                   & Ours (full) & NCM & 0.745 & \abest{0.772}{cOurs} & 0.745 \\
\bottomrule
\end{tabular}
\end{table}

\section{CONCLUSION}

This study presents a novel fMRI-based representational framework for the classification of ACAM-J states. The proposed ACAM-J-MARS approach integrates multi-atlas connectivity estimates in tangent space, low-frequency spectral power features, and within-subject alignment procedures. Using a nearest class mean decoder, our method achieves a balanced accuracy of 0.757 for binary discrimination of early versus late ACAM-J and 0.545 for three-level classification. Our results suggest that ACAM-J states are associated with distributed patterns of functional connectivity and power in slow BOLD fluctuations. However, interpretation is constrained by reliance on self-reported labels, a single-cohort dataset comprising 22 practitioners, and an alignment strategy that requires multiple runs per participant. Future work should validate the approach in independent samples and assess prospective real-time applications, including meditation-time neurofeedback derived from the proposed features.

{\small
\bibliographystyle{IEEEbib}
\bibliography{isbi2027}
}

\end{document}